\documentclass[10pt,twocolumn,letterpaper]{article}

\usepackage{cvpr}              

\usepackage{graphicx}
\usepackage{booktabs}
\usepackage{colortbl}
\usepackage{xcolor}
\usepackage{multirow}
\usepackage{amsmath}

\definecolor{cvprblue}{rgb}{0.21,0.49,0.74}
\usepackage[pagebackref,breaklinks,colorlinks,allcolors=cvprblue]{hyperref}

\def\paperID{.} 
\def\confName{.}

\title{A Novel Tuning Method for Real-time Multiple-Object Tracking Utilizing Thermal Sensor with Complexity Motion Pattern}

\author{
	Duong Nguyen-Ngoc Tran$^*$, 
	Long Hoang Pham, 
	Chi Dai Tran,
	Quoc Pham-Nam Ho, \\
	Huy-Hung Nguyen,
	Jae Wook Jeon$^\dagger$ \\
	Department of Electrical and Computer Engineering, \\
	Sungkyunkwan University, Suwon, South Korea \\
	{\tt\small $\{$duongtran, phlong, hpnquoc, tdc2000, huyhung91, jwjeon$\}$@skku.edu} 
}

\begin{document}
\maketitle

\begin{abstract}
Multi-Object Tracking in thermal images is essential for surveillance systems, particularly in challenging environments where RGB cameras struggle due to low visibility or poor lighting conditions. Thermal sensors enhance recognition tasks by capturing infrared signatures, but a major challenge is their low-level feature representation, which makes it difficult to accurately detect and track pedestrians.
To address this, the paper introduces a novel tuning method for pedestrian tracking, specifically designed to handle the complex motion patterns in thermal imagery. The proposed framework optimizes two-stages, ensuring that each stage is tuned with the most suitable hyperparameters to maximize tracking performance.
By fine-tuning hyperparameters for real-time tracking, the method achieves high accuracy without relying on complex reidentification or motion models. Extensive experiments on PBVS Thermal MOT dataset demonstrate that the approach is highly effective across various thermal camera conditions, making it a robust solution for real-world surveillance applications.
The source code is available at \url{https://github.com/DuongTran1708/pbvs25_tp-mot}
\end{abstract}

\section{Introduction}
\label{section:introduction}

\begin{figure}[t]
	\begin{center}
		\setlength{\tabcolsep}{2pt} 
		\begin{tabular}{c}
			\includegraphics[height=0.7\linewidth]{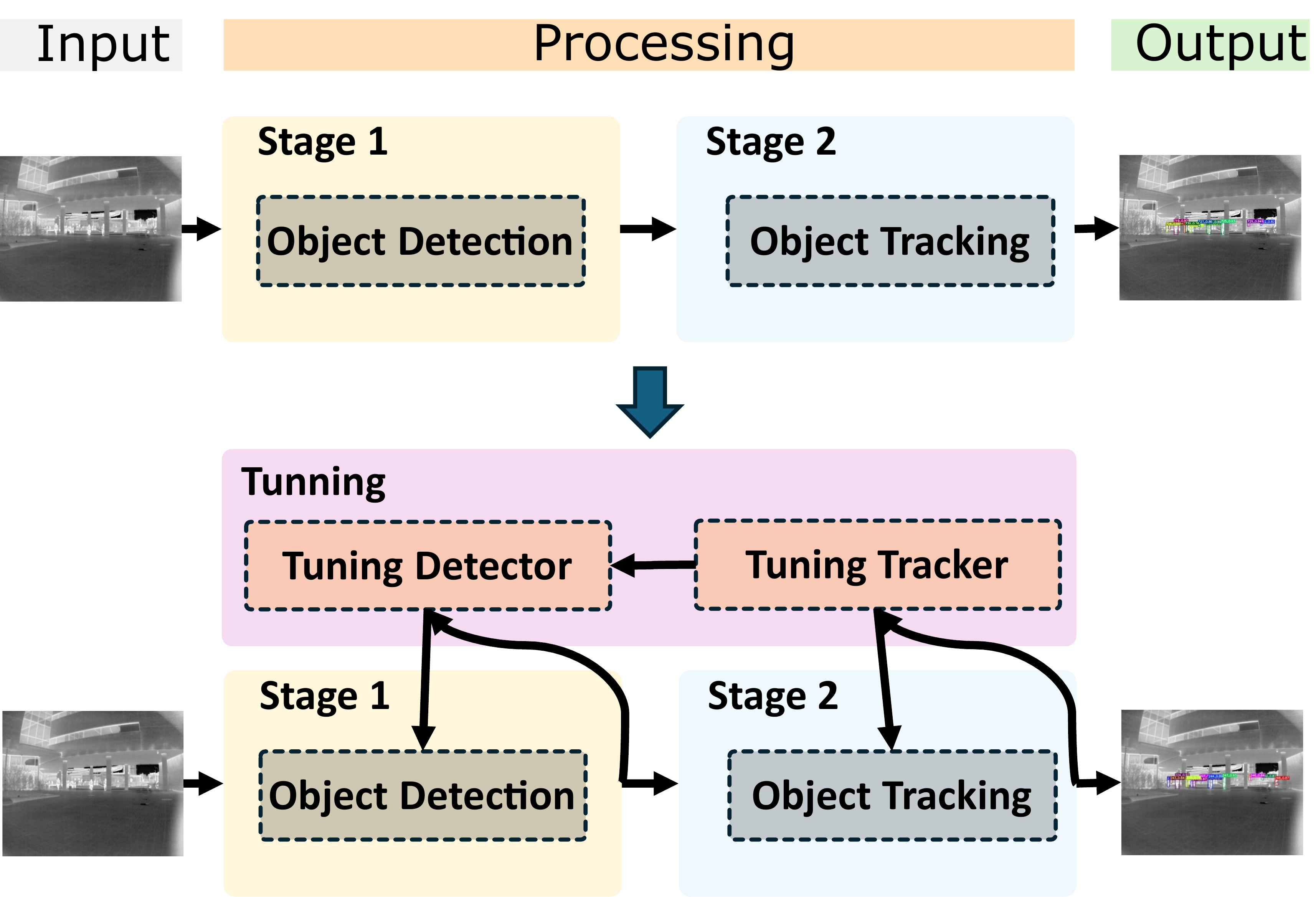} \\
		\end{tabular}
	\end{center}
	\caption{Comparison between typical two-stage multi-object tracking (upper) and our proposed framework (below). By adding the tuning, which fits each scene and optimizes the overall tracking performance, our proposed framework enhances the typical two-stage approach. The improvement in tracking accuracy is significant and demonstrates the effectiveness of our approach.}
	\label{fig:main_idea}
\end{figure}

Multi-Object Tracking (MOT) in thermal imagery has gained significant attention in recent years due to its crucial role in surveillance, security systems, and autonomous navigation. Unlike RGB cameras, which rely on visible light to detect and track objects, thermal sensors capture infrared radiation emitted by objects, making them highly effective in low-light and challenging environments. This capability is particularly beneficial in night-time surveillance, foggy conditions, or extreme weather, where traditional RGB cameras often fail.
Despite these advantages, thermal-based MOT presents several challenges, particularly in low-level feature representation. Since thermal images lack rich texture and color information, detecting and distinguishing objects—especially pedestrians—becomes more difficult. Pedestrian tracking in thermal imagery is particularly complex due to frequent occlusions, dynamic motion patterns, and low-contrast object boundaries. These factors make it difficult for conventional tracking algorithms to perform with high accuracy, necessitating advanced techniques for improving tracking performance.
The growing demand for real-time tracking solutions in security, defense, and smart city applications has further emphasized the need for robust tracking frameworks that work effectively in low-input feature environments. Developing high-accuracy multi-object tracking (MOT) models for thermal images is crucial for improving pedestrian detection, anomaly detection, and behavior analysis in various applications, including border surveillance, search and rescue missions, and automated driving systems.

While MOT has been widely explored for RGB images, thermal-based MOT remains an open challenge due to several unique difficulties.
One of the primary challenges is the low-level feature representation in thermal images. Unlike RGB images, where texture and color help distinguish objects, thermal images only capture heat signatures. This makes it difficult to differentiate between objects, especially in high-density environments or when objects have similar heat emissions. Additionally, pedestrian tracking becomes particularly problematic due to frequent occlusions, variations in body temperature, and the influence of background heat sources.
Another major challenge is the variability of environmental conditions. Thermal imagery is affected by temperature fluctuations, humidity, and background heat signatures, which can distort object appearances and introduce noise into tracking algorithms. This issue is particularly significant in outdoor surveillance and autonomous navigation, where temperature changes throughout the day impact the reliability of thermal imaging.
Furthermore, complex motion patterns add another layer of difficulty to thermal-based MOT. Pedestrians and moving objects exhibit non-linear trajectories, including sudden stops, occlusions, and rapid directional changes. Traditional motion models, such as the Kalman Filter, struggle to predict and adapt to these changes accurately, leading to tracking failures. These challenges highlight the need for robust multi-object tracking models that can effectively handle low feature resolution, motion unpredictability, and environmental variability while maintaining real-time efficiency.

To address these challenges, this paper introduces a novel hyperparameter tuning method designed specifically for multi-object tracking in thermal images. Unlike conventional approaches that rely on complex motion models and re-identification strategies, this framework focuses on fine-tuning the key parameters in two main tracking stages to achieve high accuracy with minimal computational overhead.
The core innovation of the proposed approach lies in its stage-wise hyperparameter optimization. By dividing the tracking process into two distinct stages, the framework ensures that each stage utilizes the most suitable settings, improving both detection accuracy and object association. This approach allows for more precise object tracking while reducing the need for computationally expensive re-identification techniques.
Additionally, the framework incorporates an adaptive object association mechanism, which eliminates the need for complex identity re-matching models. This makes the method lightweight and real-time, allowing it to function efficiently even in resource-constrained environments. By fine-tuning hyperparameters at each stage, the framework ensures high tracking accuracy across different environmental conditions, making it robust to occlusions, high thermal noise, and variations in pedestrian motion patterns.
Another key advantage of the proposed solution is its real-time processing capability. Unlike deep-learning-based approaches that require large-scale computations, this framework is optimized for fast execution, making it ideal for real-world deployment in surveillance systems, security cameras, and autonomous navigation. Through careful hyperparameter tuning, the framework significantly enhances tracking performance while maintaining computational efficiency.

In brief, the main contributions are as follows:
\begin{itemize}
  \item The proposed framework improves multi-object tracking in thermal imagery by introducing a two-stage tracking approach, where detection and association are separately optimized for better accuracy.
  \item A key innovation is its hyperparameter tuning mechanism, which dynamically adjusts parameters to enhance tracking performance without relying on computationally expensive re-identification (ReID) models.
  \item By prioritizing direct association tuning, the model reduces computational overhead while maintaining high tracking accuracy, making it effective for real-time applications.
  \item Designed for real-time deployment, the framework achieves high efficiency and accuracy, making it ideal for surveillance, security, and industrial monitoring systems in challenging environments.
\end{itemize}
The rest of this paper is organized as follows. Section \ref{section:related_works} discusses the related works and methods in detail. Section \ref{section:methodology} describes the proposed method. Section \ref{section:experiments_discussion} shows the optimizing process and qualitative results of the proposed method. Section \ref{section:conclusion} presents the conclusions of this work.

\section{Related Works}
\label{section:related_works}

\subsection{Object Detection}
\label{subsection:object_detection} 

Object detection, a critical task in computer vision, involves identifying and locating objects within images or videos. Recent breakthroughs in deep learning have positioned these techniques as the dominant method for object detection, delivering impressive accuracy and performance. Two primary approaches have emerged: two-stage detectors and single-stage detectors.

Two-stage detectors, such as R-CNN \cite{girshick_rich_2014}, Fast R-CNN \cite{girshick_fast_2015}, and Faster R-CNN \cite{ren_faster_2017}, operate by first generating region proposals using a separate model or algorithm, then classifying objects within those regions. While this approach offers high precision, it tends to be slower than single-stage detectors. In contrast, single-stage detectors like YOLO \cite{redmon_you_2016} and SSD \cite{leibe_ssd_2016} streamline the process by predicting object classes and bounding box coordinates in a single pass, bypassing the region proposal step. Although these models excel in speed, they often sacrifice accuracy, particularly for small or overlapping objects.

YOLO (You Only Look Once) \cite{redmon_you_2016} revolutionized object detection with its innovative approach that prioritized speed and efficiency. YOLO has the ability to detect objects in real time by processing an entire image in a single pass, in contrast to previous two-stage detectors. A more sophisticated architecture, which includes a feature pyramid network, further improved the efficacy of YOLOv3 \cite{redmon_yolov3_2018}. This architecture enables the better detection of objects at various dimensions. In order to accomplish cutting-edge outcomes, YOLOv4 \cite{bochkovskiy_yolov4_2020} implemented a "bag of freebies" and a "bag of specials," which integrated a variety of optimization techniques. The series continued to develop with the release of YOLOv5 \cite{Jocher_Ultralytics_YOLO_2023}, which marked a significant transition to a PyTorch implementation. Subsequent versions, including YOLOv6 \cite{li_yolov6_2022}, YOLOv7 \cite{wang_yolov7_2023}, and YOLOv8 \cite{Jocher_Ultralytics_YOLO_2023}, each introduced architectural refinements, training enhancements, and frequently, additional performance optimizations. The concept of a definitive YOLOv11 \cite{Jocher_Ultralytics_YOLO_2023} as a singular, officially released model is less concrete due to the continuous evolution of the YOLO family, despite the fact that YOLOv9 \cite{leonardis_yolov9_2025} represents a significant leap with innovations such as Programmable Gradient Information (PGI) and Generalized Efficient Layer Aggregation Network (GELAN). Rather, YOLOv11 \cite{Jocher_Ultralytics_YOLO_2023} can be interpreted as a representation of the YOLO framework's continuous advancement and cumulative enhancements, which are indicative of the ongoing pursuit of enhanced performance and efficiency in real-time object detection.

In this work, the author implements YOLOv8s for small-scale and real-time processing, which is appropriate for edge devices.

\subsection{Multiple Objects Tracking}
\label{subsection:multiple_objects_tracking} 

Recently, SORT \cite{bewley_simple_2016}, DeepSORT \cite{wojke_simple_2017}, and ByteTrack \cite{avidan_bytetrack_2022} have emerged as some of the most prevalent and extensively utilized approaches for multiple object tracking. SORT \cite{bewley_simple_2016} employs a tracking-by-detection methodology, linking detections from prior and current frames using data association and state estimate techniques grounded in the Kalman filter. It also facilitates object re-entry within a specified time period and manages partial occlusion. DeepSORT \cite{wojke_simple_2017} enhances SORT \cite{bewley_simple_2016} by including a deep association measure based on picture attributes. In ByteTrack \cite{avidan_bytetrack_2022}, all detections are linked regardless of their low confidence ratings, hence enhancing the efficacy of monitoring many objects in intricate surroundings.
BoT-SORT \cite{aharon_bot-sort_2022} enhances traditional SORT-like algorithms by incorporating camera motion compensation, an improved Kalman filter state vector, and a robust IoU-ReID fusion method. BoostTrack++ \cite{stanojevic_boosttrack_2024,stanojevic_boosttrack_2024_pp} introduces a soft detection confidence boost technique and refines similarity metrics using shape constraints, Mahalanobis distance, and soft BIoU similarity to improve tracking accuracy. To handle non-linear motion prediction, DiffMOT \cite{lv_diffmot_2024} is a real-time multiple object tracker introduces a diffusion probabilistic model. It employs a Decoupled Diffusion-based Motion Predictor to model complex motion patterns.

In the study, we will evaluate the majority of contemporary real-time monitoring technologies to determine which one achieves the best ranking.

\subsection{Thermal Dataset}
\label{subsection:thermal_dataset} 

The Thermal MOT Dataset from PBVS \cite{ahmar_enhancing_2024} is the first exhaustive thermal imaging dataset with annotations tailored for tracking multiple objects. It was compiled using a FLIR ADK thermal sensor, capturing 30 sequences (totaling 9,000 frames) across five urban intersections. These sequences provide a robust benchmark for thermal multi-object tracking (MOT) research, encompassing disparate public environments and object types.

The KAIST Multispectral Pedestrian Detection Benchmark \cite{hwang_multispectral_2015} is a valuable resource for advancing pedestrian detection studies, particularly in challenging conditions. This dataset consists of aligned RGB and thermal infrared image pairs, recorded from a vehicle traversing numerous traffic scenarios during both day and night. Comprising 95,000 scrupulously annotated color-thermal image pairings, it includes bounding boxes for pedestrians, cyclists, and people. With over 103,000 detailed annotations and 1,182 unique individuals, it offers a rich and varied dataset for training and evaluating pedestrian detection algorithms.

A Thermal Infrared Pedestrian Tracking Benchmark (PTB-TIR) \cite{liu_ptb-tir_2020} is a substantial dataset for advancing thermal pedestrian tracking research, encompassing over two hours of recording time across 60 distinct thermal infrared sequences. This extensive collection provides a rich foundation for algorithm development and evaluation, featuring a total of 33,745 frames, all meticulously annotated manually. The dataset's scale and the manual annotation of all sequences underscore its value as a comprehensive resource for researchers focused on pedestrian trajectory analysis and algorithm benchmarking in the thermal infrared domain.

\section{Methodology}
\label{section:methodology}

\begin{figure*}[t]
	\centering
	\includegraphics[width=1.0\textwidth]{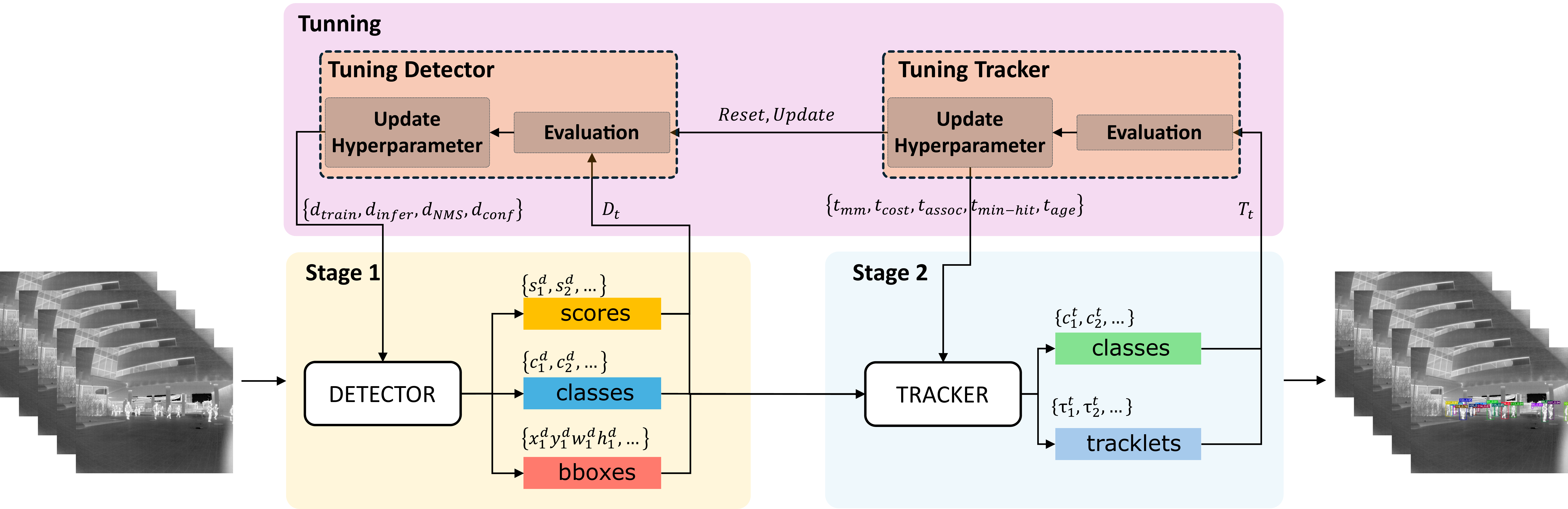}
	\caption{Using the thermal sensor input, Stage 1 involves running the detector to identify each object's class, location, and confidence score. In Stage 2, the tracker refines object locations and predicts their movements. Throughout both stages, the detector and tracker adjust hyperparameters based on evaluation results, either updating or resetting them for improved accuracy..}
	\label{fig:framework}
\end{figure*}

\subsection{Stage 1: Real-Time Object Detection}
\label{subsection:stage_1_real_time_object_detection}

The detection stage involves identifying objects (e.g., pedestrians) in each video frame using a pre-trained object detector, such as YOLO. The detector outputs bounding boxes and confidence scores for each detected object. For a given frame at time $ t $, the detector generates a set of detections:
\begin{align}
D_t = \{d_1, d_2, \dots, d_n\},
\end{align}
where each detection $ d_i $ is defined as:
\begin{align}
d_i = (x^d_i, y^d_i, w^d_i, h^d_i, s^d_i),
\end{align}
with $ (x^d_i, y^d_i) $ represents coordinates of the top-left corner of the bounding box, $ w^d_i, h^d_i $ illustrates width and height of the bounding box. $ s^d_i $ is confidence score indicating the likelihood of correct detection. The detection process can be mathematically expressed as:
\begin{align}
D_t = \text{Detector}(I_t),
\end{align}
where $ I_t $ is the input frame at time $ t $, and $ \text{Detector} $ represents the object detection model.

The key elements of the detection phase that we adjust in the article comprise:

\begin{itemize}
	 \item \textbf{Training Image Size} ($d_{train}$): This refers to the dimensions of the images used while training an object detection model. By using a variety of sizes, the model can learn to identify objects at different scales—like tiny insects or massive vehicles—making it more versatile and effective when applied to real-world situations where object sizes vary widely.
 	\item \textbf{Inference Image Size} ($d_{infer}$): This is the size of the images fed into the model when it’s actively detecting objects after training. The choice of size affects both how quickly the model processes the image and how precise its detections are. Larger images might slow things down but can improve accuracy, especially for spotting smaller objects that need more detail to be recognized.
	 \item \textbf{Non-Maximum Suppression (NMS)} ($d_{NMS}$): This is a cleanup process that happens after the model makes its initial detections. Often, the model predicts multiple overlapping boxes around a single object; NMS steps in to pick the box with the highest confidence score and discards the rest. This ensures each object is represented by just one clear bounding box, avoiding cluttered or redundant results.
 	\item \textbf{Confidence Threshold} ($d_{conf}$): This acts as a filter for the model’s predictions. Every detection comes with a confidence score indicating how certain the model is about it. The threshold is a cutoff point—detections with scores below it are ignored, which helps cut down on false positives and keeps the output reliable by only keeping predictions the model strongly believes in.
\end{itemize}

Therefore, to tune the hyperparameters of the key components we mentioned above, we update or reset them after each evaluation based on the variation:
\begin{align}
	\begin{cases}
		d^{k}_{train} &= d^{k - 1}_{train} + \Delta d_{train} \\
		d^{k}_{infer} &= d^{k - 1}_{infer} + \Delta d_{infer} \\
		d^{k}_{NMS} &= d^{k - 1}_{NMS} + \Delta d_{NMS} \\
		d^{k}_{conf} &= d^{k - 1}_{conf} + \Delta d_{conf}
	\end{cases}
\end{align}
where $k$ mean the step of each adjustment.

\subsection{Stage 2: Multi-Object Tracking}
\label{subsection:stage_2_multi_object_tracking}

After running the tracker, we obtain the object’s tracklet:
\begin{align} 
	\mathcal{T}_i=\left\{p_i^{t_1},p_i^{t_2},\ldots,p_i^{t_j}\right\}
\end{align}
\noindent where $p_i^{t_j}=\left(x_i^{t_j},y_i^{t_j}\right)$ is the coordinate point of the vehicle, $t_1$ is the time when the vehicle is become moving object, $t_j$ is the time the detection match with the current tracklet. The tracking process can be mathematically expressed as:
\begin{align}
	T_t = \text{Tracker}(I_t),
\end{align}

The association stage links detections across frames to form continuous object tracks. This stage involves predicting the next position of each tracked object and matching these predictions to new detections using a cost function.

Key components of the tracking stage we tune in the paper include:
\begin{itemize}
    \item \textbf{Motion Model} ($t_{mm}$): A Kalman Filter \cite{kalman_new_1960} predicts the next position of each object based on its previous state, including position and velocity. Additionally, diffusion-based motion models, such as the Decoupled Diffusion-Based Motion Predictor (D$^2$MP) \cite{lv_diffmot_2024}, offer an alternative approach for handling complex motion patterns in multi-object tracking
    \item \textbf{Cost Function} ($t_{cost}$): The cost function evaluates the similarity between predicted positions and new detections, commonly using Euclidean distance or Intersection over Union (IoU) to determine the best match.
    \item \textbf{Association Algorithm} ($t_{assoc}$): The Hungarian Algorithm assigns detections to tracks by minimizing the total cost of association. Additionally, appearance-based association methods, such as Re-Identification (ReID) feature matching, can be used to track objects across frames, especially in cases of occlusion or long-term tracking scenarios.
    \item \textbf{Memory Aware} ($t_{min-hit}, t_{age}$): The age of a moving object in a memory bank is determined by the number of frames it has been tracked. Additionally, a minimum number of detection hits is required to initialize a new object track, ensuring robustness against false positives and improving tracking stability.
\end{itemize}

Therfore, for tunning the hyperparameter of key components we mentione above, we update or reset after each evaluation from the valiation:
\begin{align}
	\begin{cases}
			t^{h}_{mm} &= t^{h - 1}_{mm} + \Delta t_{mm} \\
			t^{h}_{cost} &= t^{h - 1}_{cost} + \Delta t_{cost} \\
			t^{h}_{assoc} &= t^{h - 1}_{assoc} + \Delta t_{assoc} \\
			t^{h}_{min-hit} &= t^{h - 1}_{min-hit} + \Delta t_{min-hit} \\
			t^{h}_{age} &= t^{h - 1}_{age} + \Delta t_{age}
	\end{cases}
\end{align}
where $h$ mean the step of each adjustment.
\begin{table}[t]
	\centering
	\resizebox{1.0\linewidth}{!} {
		\begin{tabular}{@{}lll@{}}
			\toprule
			\multicolumn{1}{c}{\textbf{Metric}} & \multicolumn{1}{c}{\textbf{Meaning}}                      & \multicolumn{1}{c}{\textbf{Better}} \\ \midrule
			\textbf{IDF1}                       & Identity-based   F1-score (balance of IDP \& IDR)         & ↑ Higher                                  \\
			\textbf{IDP}                        & Identity Precision   (correct ID assignments)             & ↑ Higher                                  \\
			\textbf{IDR}                        & Identity Recall   (correct IDs over GT)                   & ↑ Higher                                  \\
			\textbf{Rcll}                       & Recall (true   detections / total GT)                     & ↑ Higher                                  \\
			\textbf{Prcn}                       & Precision (true   detections / all detections)            & ↑ Higher                                  \\
			\textbf{FAR}                        & False Alarm Rate (FP   per frame)                         & ↓ Lower                                   \\
			\textbf{GT}                         & Total ground truth   objects                              & --                                        \\
			\textbf{MT}                         & Mostly tracked   (tracked \textgreater{}80\% of lifetime) & ↑ Higher                                  \\
			\textbf{PT}                         & Partially tracked   (tracked 20-80\%)                     & --                                        \\
			\textbf{ML}                         & Mostly lost (tracked   \textless{}20\% of lifetime)       & ↓ Lower                                   \\
			\textbf{FP}                         & False positives   (wrong detections)                      & ↓ Lower                                   \\
			\textbf{FN}                         & False negatives   (missed objects)                        & ↓ Lower                                   \\
			\textbf{IDs}                        & Identity switches   (track ID errors)                     & ↓ Lower                                   \\
			\textbf{FM}                         & Fragmentation (broken   tracklets)                        & ↓ Lower                                   \\
			\textbf{MOTA}                       & Tracking accuracy   (overall errors)                      & ↑ Higher                                  \\
			\textbf{MOTP}                       & Tracking precision   (bounding box accuracy)              & ↓ Lower                                   \\
			\textbf{MOTAL}                      & MOTA with ID penalty                                      & ↑ Higher                                  \\ \bottomrule
		\end{tabular}
	}
	\caption{The Multi-Object Tracking metrics.}
	\label{tab:multi_object_tracking_metric}
\end{table}
\section{Experiments \& Discussion}
\label{section:experiments_discussion}

\subsection{Implementation Details}
\label{subsection:implementation_details}

The framework was implemented on a desktop system with an Intel Core i7-7700 CPU, an NVIDIA GeForce RTX 3090 (24GB VRAM), and 32GB RAM. The implementation utilizes a combination of OpenCV and PyTorch libraries, along with source code from various existing trackers to facilitate parameter tuning and evaluation. Each tracker is configured with its own custom settings, allowing for optimized performance adjustments. Additionally, for trackers with publicly available pretrained weights on GitHub, the original weights were used without additional training for re-identification, ensuring consistency with prior benchmarks.

\subsection{Evaluation Metric}
\label{subsection:evaluation_metric}

The PBVS TP-MOT challenge ranks participating teams based on three key evaluation metrics: MOTA (Multiple Object Tracking Accuracy), MOTP (Multiple Object Tracking Precision), and IDF1 (Identification F1 Score). The ranking prioritization follows a hierarchical order: MOTA is the primary ranking criterion, meaning teams are first compared based on their overall tracking accuracy.

\begin{equation}
	MOTA = 1 - \frac{FN + FP + IDSW}{GT}
	\label{eq:mota_equation}
\end{equation}

If multiple teams achieve the same MOTA score, MOTP is used as a tiebreaker, assessing the precision of the detected object locations. 

\begin{equation}
        MOTP = \frac{\sum_{i,t} d_{i,t}}{\sum_t c_t}
	\label{eq:motp_equation}
\end{equation}

If teams still have identical rankings after considering both MOTA and MOTP, IDF1 is used as the final criterion, evaluating the quality of identity preservation throughout the tracking sequence. This ranking system ensures that tracking accuracy, spatial precision, and identity consistency are all considered when determining the best-performing models in the challenge.

\begin{equation}
	IDF1 = \frac{2 \times IDP \times IDR}{IDP + IDR}
	\label{eq:idf1_equation}
\end{equation}
\begin{table*}[]
	\resizebox{1.0\textwidth}{!} {
		\begin{tabular}{@{}c|llllllllllll@{}}
			\toprule
			\cellcolor[HTML]{FFFFFF}                               & \multicolumn{6}{c}{\textbf{Average Precision}}                                                                                                                                                                                                         & \multicolumn{6}{c}{\textbf{Average Recall}}                                                                                                                                                                                                         \\ \cmidrule(l){2-13} 
			\cellcolor[HTML]{FFFFFF}                               & \multicolumn{1}{c}{\textbf{0.50:0.95}} & \multicolumn{1}{c}{\textbf{0.5}} & \multicolumn{1}{c}{\textbf{0.75}} & \multicolumn{1}{c}{\textbf{0.50:0.95}} & \multicolumn{1}{c}{\textbf{0.50:0.95}} & \multicolumn{1}{c|}{\textbf{0.50:0.95}}              & \multicolumn{1}{c}{\textbf{0.50:0.95}} & \multicolumn{1}{c}{\textbf{0.50:0.95}} & \multicolumn{1}{c}{\textbf{0.50:0.95}} & \multicolumn{1}{c}{\textbf{0.50:0.95}} & \multicolumn{1}{c}{\textbf{0.50:0.95}} & \multicolumn{1}{c}{\textbf{0.50:0.95}} \\
			\cellcolor[HTML]{FFFFFF}                               & \multicolumn{1}{c}{\textbf{all}}       & \multicolumn{1}{c}{\textbf{all}} & \multicolumn{1}{c}{\textbf{all}}  & \multicolumn{1}{c}{\textbf{small}}     & \multicolumn{1}{c}{\textbf{medium}}    & \multicolumn{1}{c|}{\textbf{large}}                  & \multicolumn{1}{c}{\textbf{all}}       & \multicolumn{1}{c}{\textbf{all}}       & \multicolumn{1}{c}{\textbf{all}}       & \multicolumn{1}{c}{\textbf{small}}     & \multicolumn{1}{c}{\textbf{medium}}    & \multicolumn{1}{c}{\textbf{large}}     \\
			\multirow{-4}{*}{\cellcolor[HTML]{FFFFFF}$\mathbf{d_{NMS}}$} & \multicolumn{1}{c}{\textbf{100}}       & \multicolumn{1}{c}{\textbf{100}} & \multicolumn{1}{c}{\textbf{100}}  & \multicolumn{1}{c}{\textbf{100}}       & \multicolumn{1}{c}{\textbf{100}}       & \multicolumn{1}{c|}{\textbf{100}}                    & \multicolumn{1}{c}{\textbf{1}}         & \multicolumn{1}{c}{\textbf{10}}        & \multicolumn{1}{c}{\textbf{100}}       & \multicolumn{1}{c}{\textbf{100}}       & \multicolumn{1}{c}{\textbf{100}}       & \multicolumn{1}{c}{\textbf{100}}       \\ \midrule
			\cellcolor[HTML]{FFFFFF}0.20                           & 0.8305                                 & 0.9461                           & 0.9004                            & 0.7943                                 & 0.9287                                 & \multicolumn{1}{l|}{-0.0336}                         & 0.1432                                 & 0.7996                                 & 0.8473                                 & 0.8172                                 & 0.9382                                 & -0.0315                                \\
			\cellcolor[HTML]{FFFFFF}0.30                           & 0.8437                                 & 0.9601                           & 0.9162                            & 0.8084                                 & 0.9368                                 & \multicolumn{1}{l|}{-0.0201}                         & 0.1432                                 & 0.8077                                 & 0.8611                                 & 0.8313                                 & 0.9464                                 & -0.0183                                \\
			\cellcolor[HTML]{FFFFFF}0.40                           & 0.8501                                 & 0.9684                           & 0.9215                            & 0.8129                                 & 0.9439                                 & \multicolumn{1}{l|}{-0.0167}                         & 0.1432                                 & 0.8116                                 & 0.8678                                 & 0.8366                                 & 0.9536                                 & -0.0151                                \\
			\cellcolor[HTML]{FFFFFF}0.50                           & 0.8549                                 & 0.9731                           & 0.9288                            & 0.8175                                 & 0.9483                                 & \multicolumn{1}{l|}{-0.0162}                         & 0.1432                                 & 0.8145                                 & 0.8730                                 & 0.8417                                 & 0.9584                                 & -0.0145                                \\
			\cellcolor[HTML]{FFFFFF}0.60                           & 0.8595                                 & 0.9762                           & 0.9331                            & 0.8196                                 & 0.9542                                 & \multicolumn{1}{l|}{-0.0140}                         & 0.1432                                 & 0.8180                                 & 0.8774                                 & 0.8442                                 & 0.9642                                 & -0.0125                                \\
			\rowcolor[HTML]{FFFFFF} 
			0.70                                                   & 0.8616                                 & 0.9789                           & 0.9367                            & 0.8216                                 & 0.9552                                 & \multicolumn{1}{l|}{\cellcolor[HTML]{FFFFFF}-0.0093} & 0.1432                                 & 0.8200                                 & 0.8802                                 & 0.8469                                 & 0.9652                                 & -0.0078                                \\
			\rowcolor[HTML]{D9D9D9} 
			0.75                                                   & 0.8620                                 & 0.9787                           & 0.9372                            & 0.8223                                 & 0.9553                                 & \multicolumn{1}{l|}{\cellcolor[HTML]{D9D9D9}-0.0051} & 0.1432                                 & 0.8203                                 & 0.8815                                 & 0.8481                                 & 0.9656                                 & -0.0037                                \\
			\rowcolor[HTML]{FFFFFF} 
			0.80                                                   & 0.8628                                 & 0.9783                           & 0.9378                            & 0.8229                                 & 0.9553                                 & \multicolumn{1}{l|}{\cellcolor[HTML]{FFFFFF}-0.0040} & 0.1432                                 & 0.8210                                 & 0.8828                                 & 0.8500                                 & 0.9654                                 & -0.0033                                \\
			\rowcolor[HTML]{FFFFFF} 
			0.90                                                   & 0.8640                                 & 0.9746                           & 0.9388                            & 0.8239                                 & 0.9558                                 & \multicolumn{1}{l|}{\cellcolor[HTML]{FFFFFF}-0.0036} & 0.1432                                 & 0.8225                                 & 0.8883                                 & 0.8575                                 & 0.9659                                 & -0.0029                                \\
			\cellcolor[HTML]{FFFFFF}0.95                           & 0.8498                                 & 0.9487                           & 0.9211                            & 0.8005                                 & 0.9569                                 & \multicolumn{1}{l|}{-0.0040}                         & 0.1432                                 & 0.8172                                 & 0.8951                                 & 0.8662                                 & 0.9680                                 & -0.0019                                \\
			\cellcolor[HTML]{FFFFFF}1.00                           & 0.1557                                 & 0.1707                           & 0.1651                            & 0.1618                                 & 0.2323                                 & \multicolumn{1}{l|}{-0.2968}                         & 0.1432                                 & 0.2853                                 & 0.8869                                 & 0.8570                                 & 0.9746                                 & -0.0011                                \\ \bottomrule
		\end{tabular}
	}
	\caption{The result of tunning in NMS.}
	\label{tab:nms_tunning}
	\vspace*{-0.3cm}
\end{table*}
\begin{table*}[]
	\resizebox{1.0\textwidth}{!} {
		\begin{tabular}{@{}c|llllll|llllll@{}}
			\toprule
			\cellcolor[HTML]{FFFFFF}                                     & \multicolumn{6}{c|}{\textbf{Average Precision}}                                                                                                                                                                                           & \multicolumn{6}{c}{\textbf{Average Recall}}                                                                                                                                                                                                         \\ \cmidrule(l){2-13} 
			\cellcolor[HTML]{FFFFFF}                                     & \multicolumn{1}{c}{\textbf{0.50:0.95}} & \multicolumn{1}{c}{\textbf{0.5}} & \multicolumn{1}{c}{\textbf{0.75}} & \multicolumn{1}{c}{\textbf{0.50:0.95}} & \multicolumn{1}{c}{\textbf{0.50:0.95}} & \multicolumn{1}{c|}{\textbf{0.50:0.95}} & \multicolumn{1}{c}{\textbf{0.50:0.95}} & \multicolumn{1}{c}{\textbf{0.50:0.95}} & \multicolumn{1}{c}{\textbf{0.50:0.95}} & \multicolumn{1}{c}{\textbf{0.50:0.95}} & \multicolumn{1}{c}{\textbf{0.50:0.95}} & \multicolumn{1}{c}{\textbf{0.50:0.95}} \\
			\cellcolor[HTML]{FFFFFF}                                     & \multicolumn{1}{c}{\textbf{all}}       & \multicolumn{1}{c}{\textbf{all}} & \multicolumn{1}{c}{\textbf{all}}  & \multicolumn{1}{c}{\textbf{small}}     & \multicolumn{1}{c}{\textbf{medium}}    & \multicolumn{1}{c|}{\textbf{large}}     & \multicolumn{1}{c}{\textbf{all}}       & \multicolumn{1}{c}{\textbf{all}}       & \multicolumn{1}{c}{\textbf{all}}       & \multicolumn{1}{c}{\textbf{small}}     & \multicolumn{1}{c}{\textbf{medium}}    & \multicolumn{1}{c}{\textbf{large}}     \\
			\multirow{-4}{*}{\cellcolor[HTML]{FFFFFF}$\mathbf{d_{conf}}$} & \multicolumn{1}{c}{\textbf{100}}       & \multicolumn{1}{c}{\textbf{100}} & \multicolumn{1}{c}{\textbf{100}}  & \multicolumn{1}{c}{\textbf{100}}       & \multicolumn{1}{c}{\textbf{100}}       & \multicolumn{1}{c|}{\textbf{100}}       & \multicolumn{1}{c}{\textbf{1}}         & \multicolumn{1}{c}{\textbf{10}}        & \multicolumn{1}{c}{\textbf{100}}       & \multicolumn{1}{c}{\textbf{100}}       & \multicolumn{1}{c}{\textbf{100}}       & \multicolumn{1}{c}{\textbf{100}}       \\ \midrule
			\rowcolor[HTML]{D9D9D9} 
			0.0001                                                       & 0.8620                                 & 0.9787                           & 0.9372                            & 0.8223                                 & 0.9553                                 & -0.0051                                 & 0.1432                                 & 0.8203                                 & 0.8815                                 & 0.8481                                 & 0.9656                                 & -0.0037                                \\
			\cellcolor[HTML]{FFFFFF}0.001                                & 0.8614                                 & 0.9782                           & 0.9362                            & 0.8212                                 & 0.9551                                 & -0.0093                                 & 0.1432                                 & 0.8196                                 & 0.8792                                 & 0.8454                                 & 0.9652                                 & -0.0078                                \\
			\cellcolor[HTML]{FFFFFF}0.01                                 & 0.8608                                 & 0.9771                           & 0.9362                            & 0.8206                                 & 0.9551                                 & -0.0093                                 & 0.1432                                 & 0.8191                                 & 0.8783                                 & 0.8443                                 & 0.9648                                 & -0.0078                                \\
			\cellcolor[HTML]{FFFFFF}0.1                                  & 0.8590                                 & 0.9742                           & 0.9334                            & 0.8187                                 & 0.9544                                 & -0.0106                                 & 0.1432                                 & 0.8178                                 & 0.8763                                 & 0.8416                                 & 0.9643                                 & -0.0094                                \\
			\cellcolor[HTML]{FFFFFF}0.2                                  & 0.8570                                 & 0.9711                           & 0.9334                            & 0.8161                                 & 0.9539                                 & -0.0106                                 & 0.1432                                 & 0.8161                                 & 0.8742                                 & 0.8390                                 & 0.9638                                 & -0.0094                                \\
			\cellcolor[HTML]{FFFFFF}0.3                                  & 0.8556                                 & 0.9663                           & 0.9320                            & 0.8135                                 & 0.9539                                 & -0.0106                                 & 0.1432                                 & 0.8145                                 & 0.8722                                 & 0.8363                                 & 0.9638                                 & -0.0094                                \\
			\cellcolor[HTML]{FFFFFF}0.4                                  & 0.8531                                 & 0.9598                           & 0.9289                            & 0.8105                                 & 0.9539                                 & -0.0106                                 & 0.1432                                 & 0.8128                                 & 0.8700                                 & 0.8334                                 & 0.9638                                 & -0.0094                                \\
			\cellcolor[HTML]{FFFFFF}0.5                                  & 0.8483                                 & 0.9534                           & 0.9226                            & 0.8035                                 & 0.9528                                 & -0.0106                                 & 0.1432                                 & 0.8087                                 & 0.8652                                 & 0.8264                                 & 0.9633                                 & -0.0094                                \\
			\cellcolor[HTML]{FFFFFF}0.6                                  & 0.8422                                 & 0.9386                           & 0.9194                            & 0.7954                                 & 0.9528                                 & -0.0106                                 & 0.1432                                 & 0.8045                                 & 0.8594                                 & 0.8174                                 & 0.9633                                 & -0.0094                                \\
			\cellcolor[HTML]{FFFFFF}0.7                                  & 0.8200                                 & 0.9074                           & 0.8909                            & 0.7609                                 & 0.9523                                 & -0.0106                                 & 0.1429                                 & 0.7841                                 & 0.8344                                 & 0.7812                                 & 0.9629                                 & -0.0094                                \\
			\cellcolor[HTML]{FFFFFF}0.8                                  & 0.7700                                 & 0.8372                           & 0.8299                            & 0.6909                                 & 0.9508                                 & -0.0106                                 & 0.1374                                 & 0.7459                                 & 0.7833                                 & 0.7095                                 & 0.9611                                 & -0.0094                                \\ \bottomrule
		\end{tabular}
	}
	\caption{The result of tunning in confindent threshold.}
	\label{tab:threshold_tunning}
	\vspace*{-0.25cm}
\end{table*}
\begin{table*}[]
	\begin{center}	
		\resizebox{1.0\textwidth}{!} {
			\begin{tabular}{@{}ccccccccc|cccccc|cccccc@{}}
				\toprule
				\multicolumn{4}{c}{}                                                                         & \multicolumn{5}{c|}{}                                                       & \multicolumn{6}{c|}{\textbf{Average Precision}}                                                                  & \multicolumn{6}{c}{\textbf{Average Recall}}                                                                                 \\ \cmidrule(l){10-21} 
				\multicolumn{4}{c}{}                                                                         & \multicolumn{5}{c|}{}                                                       & \textbf{0.50:0.95} & \textbf{0.5} & \textbf{0.75} & \textbf{0.50:0.95} & \textbf{0.50:0.95} & \textbf{0.50:0.95} & \textbf{0.50:0.95} & \textbf{0.50:0.95} & \textbf{0.50:0.95} & \textbf{0.50:0.95} & \textbf{0.50:0.95} & \textbf{0.50:0.95} \\
				\multicolumn{4}{c}{\multirow{-3}{*}{$\mathbf{d_{train}}$}}                          & \multicolumn{5}{c|}{\multirow{-3}{*}{$\mathbf{d_{infer}}$}}      & \textbf{all}       & \textbf{all} & \textbf{all}  & \textbf{small}     & \textbf{medium}    & \textbf{large}     & \textbf{all}       & \textbf{all}       & \textbf{all}       & \textbf{small}     & \textbf{medium}    & \textbf{large}     \\ \cmidrule(r){1-9}
				\textbf{640} & \textbf{960} & \textbf{1280} & \multicolumn{1}{c|}{\textbf{1600}}             & \textbf{640} & \textbf{960} & \textbf{1280} & \textbf{1600} & \textbf{1760} & \textbf{100}       & \textbf{100} & \textbf{100}  & \textbf{100}       & \textbf{100}       & \textbf{100}       & \textbf{1}         & \textbf{10}        & \textbf{100}       & \textbf{100}       & \textbf{100}       & \textbf{100}       \\ \midrule
				x            &              &               & \multicolumn{1}{c|}{}                          & x            &              &               &               &               & 0.8458             & 0.9545       & 0.9216        & 0.8121             & 0.9504             & -0.0066            & 0.1428             & 0.8098             & 0.8809             & 0.8481             & 0.9639             & -0.0037            \\
				& x            &               & \multicolumn{1}{c|}{}                          &              & x            &               &               &               & 0.8349             & 0.9516       & 0.9151        & 0.7858             & 0.9452             & -0.0094            & 0.1422             & 0.8057             & 0.8784             & 0.8457             & 0.9597             & -0.0035            \\
				&              & x             & \multicolumn{1}{c|}{}                          &              &              & x             &               &               & 0.8326             & 0.9524       & 0.9139        & 0.7782             & 0.9421             & -0.0073            & 0.1420             & 0.8024             & 0.8770             & 0.8448             & 0.9576             & -0.0041            \\
				\rowcolor[HTML]{D9D9D9} 
				&              &               & \multicolumn{1}{c|}{\cellcolor[HTML]{D9D9D9}x} &              &              &               & x             &               & 0.8620             & 0.9787       & 0.9372        & 0.8223             & 0.9553             & -0.0051            & 0.1432             & 0.8203             & 0.8815             & 0.8481             & 0.9656             & -0.0037            \\
				&              &               & \multicolumn{1}{c|}{x}                         &              &              &               &               & x             & 0.8598             & 0.9798       & 0.9304        & 0.8215             & 0.9511             & -0.0110            & 0.1437             & 0.8198             & 0.8817             & 0.8499             & 0.9619             & -0.0085            \\
				&              & x             & \multicolumn{1}{c|}{x}                         &              &              &               & x             &               & 0.8173             & 0.9392       & 0.9004        & 0.7603             & 0.9351             & -0.0092            & 0.1413             & 0.7937             & 0.8753             & 0.8438             & 0.9562             & -0.0043            \\
				& x            & x             & \multicolumn{1}{c|}{x}                         &              &              &               & x             &               & 0.8114             & 0.9334       & 0.8945        & 0.7542             & 0.9341             & -0.0107            & 0.1412             & 0.7911             & 0.8750             & 0.8431             & 0.9563             & -0.0049            \\
				x            & x            & x             & \multicolumn{1}{c|}{x}                         &              &              &               & x             &               & 0.8055             & 0.9256       & 0.8883        & 0.7519             & 0.9316             & -0.0106            & 0.1412             & 0.7861             & 0.8743             & 0.8422             & 0.9539             & -0.0037            \\ \bottomrule
			\end{tabular}
		}
	\end{center}
	\vspace*{-0.3cm}
	\caption{The result of tunning in image size of training and inference.}
	\label{tab:image_size_tunning}
\end{table*}
\begin{table}[]
	\begin{center}	
		\resizebox{1.\linewidth}{!} {
			\begin{tabular}{@{}c|ccccccccccccccccc@{}}
				\toprule
				\multicolumn{1}{c|}{}                               & \multicolumn{10}{c}{\textbf{Evaluation   - Onsite}}                                                                                                                                                                                                                                                                                                                                                          \\ \cmidrule(l){2-11} 
				\multicolumn{1}{c|}{\multirow{-2}{*}{$\mathbf{t_{age}}$}} & \textbf{IDF1} & \textbf{IDP} & \multicolumn{1}{c|}{\textbf{IDR}}                & \textbf{FP} & \textbf{FN} & \textbf{IDs} & \multicolumn{1}{c|}{\textbf{FM}}                & \textbf{MOTA} & \textbf{MOTP} & \textbf{MOTAL} \\ \midrule
				5                                                   & 72.8          & 73.8         & \multicolumn{1}{c|}{71.8}                          & 29          & 91          & 17           & \multicolumn{1}{c|}{53}                         & 93.7          & 86.4          & 94.4           \\
				10                                                  & 77.8          & 78.9         & \multicolumn{1}{c|}{76.7}                         & 29          & 91          & 16           & \multicolumn{1}{c|}{53}                         & 93.7          & 86.4          & 94.4           \\
				20                                                  & 80.7          & 81.9         & \multicolumn{1}{c|}{79.6}                         & 29          & 91          & 15           & \multicolumn{1}{c|}{53}                         & 93.8          & 86.4          & 94.4           \\
				30                                                  & 81.7          & 82.9         & \multicolumn{1}{c|}{80.6}                         & 29          & 91          & 15           & \multicolumn{1}{c|}{53}                         & 93.8          & 86.4          & 94.4           \\
				\rowcolor[HTML]{D9D9D9} 
				40                                                  & 82.3          & 83.5         & \multicolumn{1}{c|}{\cellcolor[HTML]{D9D9D9}81.2} & 29          & 91          & 14           & \multicolumn{1}{c|}{\cellcolor[HTML]{D9D9D9}53} & 93.8          & 86.4          & 94.4           \\
				50                                                  & 82.1          & 83.2         & \multicolumn{1}{c|}{80.9}                         & 29          & 91          & 15           & \multicolumn{1}{c|}{53}                         & 93.7          & 86.4          & 94.4           \\
				60                                                  & 81.5          & 82.7         & \multicolumn{1}{c|}{80.4}                         & 29          & 91          & 15           & \multicolumn{1}{c|}{53}                         & 93.7          & 86.4          & 94.4           \\
				70                                                  & 81.4          & 82.6         & \multicolumn{1}{c|}{80.3}                         & 29          & 91          & 16           & \multicolumn{1}{c|}{53}                         & 93.7          & 86.4          & 94.4           \\ \bottomrule
			\end{tabular}
		}
	\end{center}
	\vspace*{-0.3cm}
	\caption{The result of tunning in age of tracklet.}
	\label{tab:age_tunning}
\end{table}
\begin{table}[]
	\begin{center}	
		\resizebox{1.\linewidth}{!} {
			\begin{tabular}{@{}ccccccccccc@{}}
				\toprule
				\multicolumn{1}{c|}{}                                   & \multicolumn{10}{c}{\textbf{Evaluation   - Onsite}}                                                                                                                                                                                                                                                                                                                                                          \\ \cmidrule(l){2-11} 
				\multicolumn{1}{c|}{\multirow{-2}{*}{$\mathbf{t_{min-hit}}$}} & \textbf{IDF1} & \textbf{IDP} & \multicolumn{1}{c|}{\textbf{IDR}}                 & \textbf{FP} & \textbf{FN} & \textbf{IDs} & \multicolumn{1}{c|}{\textbf{FM}}                & \textbf{MOTA} & \textbf{MOTP} & \textbf{MOTAL} \\ \midrule
				\multicolumn{1}{c|}{1}                                                       & 80.4          & 80.8         & 79.9                                             & 85          & 116         & 14           & 54                                              & 90.1          & 85.7          & 90.6           \\
				\rowcolor[HTML]{D9D9D9} 
				\multicolumn{1}{c|}{\cellcolor[HTML]{D9D9D9}3}          & 82.3          & 83.5         & \multicolumn{1}{c|}{\cellcolor[HTML]{D9D9D9}81.2} & 29          & 91          & 14           & \multicolumn{1}{c|}{\cellcolor[HTML]{D9D9D9}53} & 93.8          & 86.4          & 94.4           \\
				\multicolumn{1}{c|}{5}                                  & 80.8          & 81.2         & \multicolumn{1}{c|}{80.4}                         & 88          & 116         & 12           & \multicolumn{1}{c|}{56}                         & 89.9          & 85.3          & 90.4           \\ \bottomrule
			\end{tabular}
		}
	\end{center}
	\vspace*{-0.3cm}
	\caption{The result of tunning in age of tracklet.}
	\label{tab:min_hit_tunning}
\end{table}
\begin{table}[]
	\begin{center}	
		\resizebox{1.\linewidth}{!} {
			\begin{tabular}{@{}c|ccccccccccccccccc@{}}
				\toprule
				& \multicolumn{10}{c}{\textbf{Evaluation   - Onsite}}                                                                                                                                                                                                                                                                                                                                                          \\ \cmidrule(l){2-11} 
				\multirow{-2}{*}{$\mathbf{t_{cost}}$} & \textbf{IDF1} & \textbf{IDP} & \multicolumn{1}{c|}{\textbf{IDR}}                & \textbf{FP} & \textbf{FN} & \textbf{IDs} & \multicolumn{1}{c|}{\textbf{FM}}                & \textbf{MOTA} & \textbf{MOTP} & \textbf{MOTAL} \\ \midrule
				\rowcolor[HTML]{D9D9D9} 
				0.01                                & 82.3          & 83.5         & \multicolumn{1}{c|}{\cellcolor[HTML]{D9D9D9}81.2} & 29          & 91          & 14           & \multicolumn{1}{c|}{\cellcolor[HTML]{D9D9D9}53} & 93.8          & 86.4          & 94.4           \\
				0.05                                & 81.6          & 82.8         & \multicolumn{1}{c|}{80.5}                        & 29          & 91          & 16           & \multicolumn{1}{c|}{53}                         & 93.7          & 86.4          & 94.4           \\
				0.1                                 & 80.6          & 81.7         & \multicolumn{1}{c|}{79.5}                         & 29          & 91          & 20           & \multicolumn{1}{c|}{53}                         & 93.5          & 86.4          & 94.4           \\
				0.2                                 & 76.1          & 77.2         & \multicolumn{1}{c|}{75.1}                         & 28          & 90          & 27           & \multicolumn{1}{c|}{53}                         & 93.2          & 86.4          & 94.4           \\ \bottomrule
			\end{tabular}
		}
	\end{center}
	\vspace*{-0.3cm}
	\caption{The result of tunning in age of tracklet.}
	\label{tab:iou_cost_tunning}
\end{table}
Moreover, the we have to test the result of tracking based on the metrics which are shown in Table \ref{tab:multi_object_tracking_metric}.

\begin{table*}[t]
	\centering
	\resizebox{1.0\textwidth}{!} {
		\begin{tabular}{@{}lccccccccccccccccc|ccccccc@{}}
			\toprule
			\multicolumn{1}{c}{} & \multicolumn{17}{c|}{\textbf{Evaluation   - Onsite}} & \multicolumn{7}{c}{\textbf{Evaluation   Server}} \\ \cmidrule(l){2-25} 
			\multicolumn{1}{c}{\multirow{-2}{*}{\textbf{Method}}} & \textbf{IDF1} & \textbf{IDP} & \multicolumn{1}{c|}{\textbf{IDR}} & \textbf{Rcll} & \textbf{Prcn} & \multicolumn{1}{c|}{\textbf{FAR}} & \textbf{GT} & \textbf{MT} & \textbf{PT} & \multicolumn{1}{c|}{\textbf{ML}} & \textbf{FP} & \textbf{FN} & \textbf{IDs} & \multicolumn{1}{c|}{\textbf{FM}} & \textbf{MOTA} & \textbf{MOTP} & \textbf{MOTAL} & \textbf{MOTA} & \multicolumn{1}{c|}{\textbf{MOTP}} & \textbf{IDF1} & \textbf{IDP} & \multicolumn{1}{c|}{\textbf{IDR}} & \textbf{RCLL} & \textbf{PRCN} \\ \midrule
			\rowcolor[HTML]{EFEFEF} 
			\multicolumn{1}{l|}{\cellcolor[HTML]{EFEFEF}SORT} & 84.3 & 84.3 & \multicolumn{1}{c|}{\cellcolor[HTML]{EFEFEF}84.2} & 99.5 & 99.6 & \multicolumn{1}{c|}{\cellcolor[HTML]{EFEFEF}0.03} & 21 & 21 & 0 & \multicolumn{1}{c|}{\cellcolor[HTML]{EFEFEF}0} & 8 & 10 & 7 & \multicolumn{1}{c|}{\cellcolor[HTML]{EFEFEF}47} & 98.8 & 87.3 & 99.1 & \textbf{98.4357} & \multicolumn{1}{c|}{\cellcolor[HTML]{EFEFEF}\textbf{0.1263}} & \textbf{81.3010} & 81.3521 & \multicolumn{1}{c|}{\cellcolor[HTML]{EFEFEF}81.2500} & 99.5499 & 99.6749 \\
			\multicolumn{1}{l|}{ByteTrack} & 77.8 & 79.2 & \multicolumn{1}{c|}{76.4} & 94.3 & 97.8 & \multicolumn{1}{c|}{0.15} & 21 & 20 & 0 & \multicolumn{1}{c|}{1} & 45 & 120 & 26 & \multicolumn{1}{c|}{58} & 91 & 86.3 & 92.1 & 91.7355 & \multicolumn{1}{c|}{0.1367} & 76.5911 & 77.7829 & \multicolumn{1}{c|}{75.4354} & 95.0118 & 97.9685 \\
			\multicolumn{1}{l|}{BoTTrack} & 77.5 & 78.7 & \multicolumn{1}{c|}{76.2} & 94.5 & 97.6 & \multicolumn{1}{c|}{0.17} & 21 & 20 & 0 & \multicolumn{1}{c|}{1} & 50 & 117 & 24 & \multicolumn{1}{c|}{56} & 91.1 & 86.3 & 92.1 & 91.7429 & \multicolumn{1}{c|}{0.1368} & 76.0549 & 77.0541 & \multicolumn{1}{c|}{75.0812} & 95.1741 & 97.6751 \\
			\multicolumn{1}{l|}{BoostTrack} & 77.3 & 80.3 & \multicolumn{1}{c|}{74.6} & 90 & 96.8 & \multicolumn{1}{c|}{0.21} & 21 & 18 & 3 & \multicolumn{1}{c|}{0} & 61 & 221 & 21 & \multicolumn{1}{c|}{57} & 86.1 & 83.9 & 86.9 & 86.5481 & \multicolumn{1}{c|}{0.1555} & 75.4542 & 78.3337 & \multicolumn{1}{c|}{72.7789} & 90.2081 & 97.0932 \\
			\multicolumn{1}{l|}{DiffMOT} & 80.2 & 83 & \multicolumn{1}{c|}{77.6} & 89.9 & 96.3 & \multicolumn{1}{c|}{0.24} & 21 & 18 & 2 & \multicolumn{1}{c|}{0} & 69 & 221 & 11 & \multicolumn{1}{c|}{61} & 85.9 & 83.3 & 86.4 & 86.3046 & \multicolumn{1}{c|}{0.1611} & 78.1231 & 80.9415 & \multicolumn{1}{c|}{75.4944} & 90.2081 & 96.7168 \\ \bottomrule
		\end{tabular}
	}
	\caption{Evaliation Result on the 1st Thermal Pedestrian Multiple Object Tracking Challenge (TP-MOT).}
	\label{tab:evaluation_result_server}
\end{table*}

\subsection{Preprocess Dataset}
\label{subsection:prepsocess_dataset} 

For the PBVS Thermal MOT dataset, we processed and renamed each image file to ensure they were in the correct sequential order. This step was necessary because the frame loader struggled to sort the images correctly, causing potential misalignment in object tracking. By renaming the files systematically, we ensured that the frames were loaded in the correct temporal sequence, improving the dataset's stability for training and evaluation.

Since thermal images are low in feature contrast and primarily grayscale, we applied several augmentations to enhance the model's robustness. The augmentations used include Fliplr (horizontal flip) to introduce variation, Crop to improve localization, Pad to maintain uniform image size, and PiecewiseAffine to apply small deformations, simulating real-world variations. These augmentations help the model generalize better, improving tracking accuracy in thermal imaging scenarios.

\subsection{Model Training}
\label{subsection:model_training} 

As can be seen in PBVS's thermal multi-object tracking dataset, the perspective of the capture sensor is the same as the person's. Therefore, in addition to the main dataset, we can use two more datasets, such as the KAIST Multispectral Pedestrian Detection Benchmark \cite{hwang_multispectral_2015} and PTB-TIR: A Thermal Infrared Pedestrian Tracking Benchmark \cite{liu_ptb-tir_2020} to improve the detection result of the detection model. 

According to the official rules of the 1st PBVS TP-MOT challenge, we are restricted to using YOLOv8 in its small version (YOLOv8s), which has an architecture equivalent to YOLOv5s. To optimize performance, we conduct extensive hyperparameter tuning and multi-scale training, experimenting with different input resolutions of 640, 960, 1280, and 1600 pixels. Training on multiple resolutions allows the model to better detect objects of varying sizes, improving detection robustness across different scenarios in the PBVS Thermal MOT dataset. 

\subsection{Tuning Experiment}
\label{subsection:tunning_experiment} 

The primary tracker optimized in this study is the Simple Online and Real-Time Tracking (SORT) algorithm, a widely adopted framework for multi-object tracking (MOT) due to its efficiency and effectiveness in real-time applications. SORT operates by integrating object detection with a two-stage tracking process: detection and association, leveraging a Kalman filter for motion prediction and the Hungarian algorithm for data association.

The default of hyperparameter are $d_{train}=1600$, $d_{infer}=1600$, $d_{nms}=0.75$, $d_{conf}=0.0001$, $t_{age}=40$, $t_{min-hit}=3$, $t_{cost}=0.01$, $t_{mm}=Kalman Filter$, and $t_{assoc}=Hungarian$.

To optimize SORT’s performance, we systematically tuned these hyperparameters, evaluating their impact on tracking accuracy and robustness. The results of this tuning process are detailed in the following tables:
\begin{itemize}
	\item Table \ref{tab:nms_tunning}: Explores variations in \( d_{nms} \) to determine the optimal NMS threshold for minimizing redundant detections while preserving true positives.
	\item Table \ref{tab:threshold_tunning}: Assesses adjustments to \( d_{conf} \), identifying the threshold that balances sensitivity and specificity in detection filtering.
	\item Table \ref{tab:image_size_tunning}: Investigates the effects of \( d_{train} \) and \( d_{infer} \) on detection and tracking performance, aiming to optimize image resolution for both training and inference phases.
	\item Table \ref{tab:age_tunning}: Analyzes \( t_{age} \) to find the ideal track lifespan, ensuring robustness against temporary occlusions without retaining stale tracks.
	\item Table \ref{tab:min_hit_tunning}: Examines \( t_{min-hit} \) to establish the minimum hits needed for reliable track initiation, reducing false track creation.
	\item Table \ref{tab:iou_cost_tunning}: Evaluates \( t_{cost} \) to pinpoint the best cost threshold for effective association, minimizing identity switches and track fragmentation.
\end{itemize}
Each table presents the tuning outcomes, identifying the best hyperparameter values that enhance the SORT framework for our specific dataset and application context.

As presented in Table \ref{tab:evaluation_result_server}, we extended our evaluation beyond the default SORT configuration to explore alternative motion models and association strategies:
\begin{itemize}
	\item For \( t_{mm} = \text{Diffusion-based} \), we consider DiffMOT \cite{lv_diffmot_2024}, a recent approach leveraging diffusion models to improve motion prediction in complex scenarios where linear assumptions of the Kalman filter may falter. This substitution aims to enhance tracking accuracy by modeling non-linear motion patterns more effectively.
	\item For \( t_{assoc} = \text{re-id} \), we evaluate BoostTrack and BoTTrack, which incorporate re-identification (re-ID) mechanisms. These methods augment the association process with appearance features derived from deep learning, improving robustness against occlusions and similar-looking objects compared to the default Hungarian algorithm’s reliance on spatial proximity alone.
\end{itemize}
These alternative configurations were tested to assess their potential to outperform the baseline SORT setup, particularly in challenging environments with frequent occlusions or erratic object movements. The results, detailed in Table \ref{tab:evaluation_result_server}, provide insights into the trade-offs between computational complexity and tracking performance, guiding the selection of optimal strategies for specific use cases such as surveillance or autonomous navigation.

\begin{figure*}[t]
	\begin{center}
		\hspace*{-1cm}
		\resizebox{1.\textwidth}{!}{
			\setlength{\tabcolsep}{2pt} 
			\renewcommand{\arraystretch}{1} 
			\begin{tabular}{cccccc}
				\textbf{seq2} &
				\textbf{seq17} &
				\textbf{seq22} &
				\textbf{seq47} &
				\textbf{seq54} &
				\textbf{seq66} \\
				\includegraphics[width=0.22\linewidth]{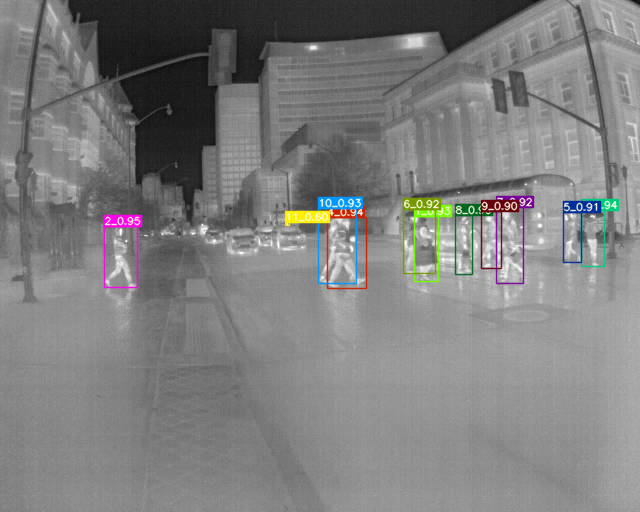} &
				\includegraphics[width=0.22\linewidth]{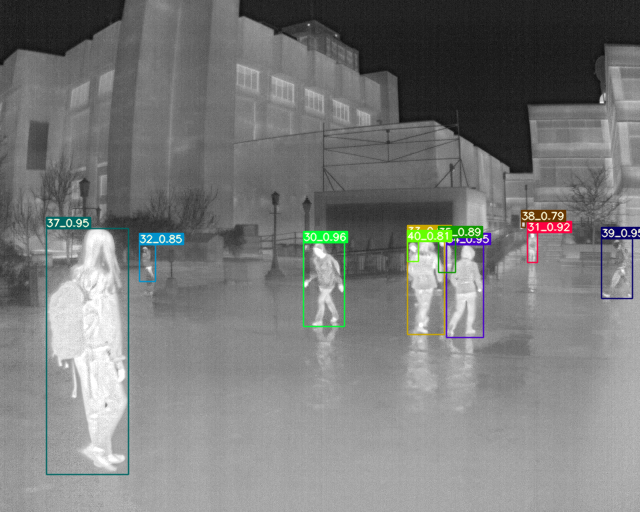} &
				\includegraphics[width=0.22\linewidth]{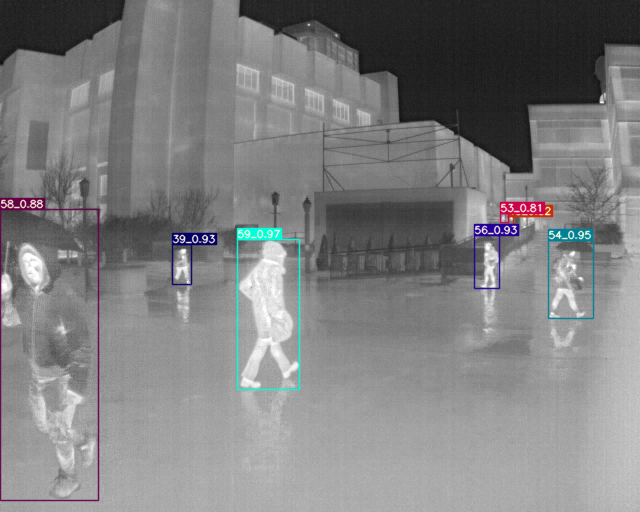} &
				\includegraphics[width=0.22\linewidth]{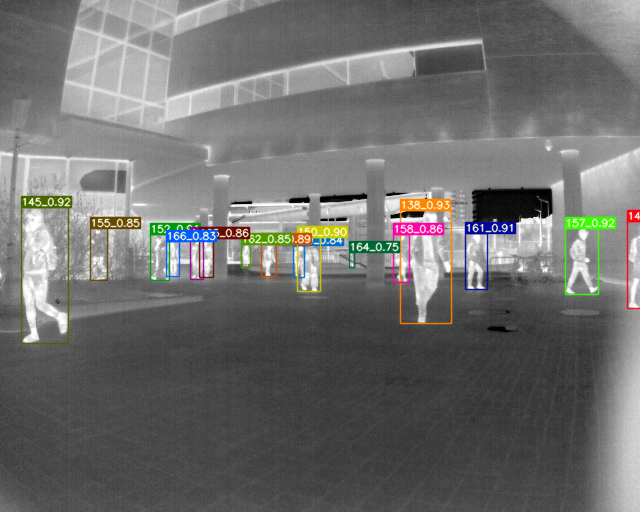} &
				\includegraphics[width=0.22\linewidth]{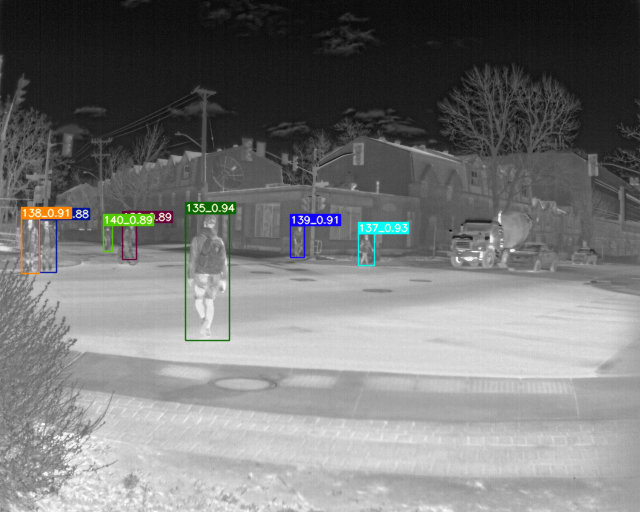} &
				\includegraphics[width=0.22\linewidth]{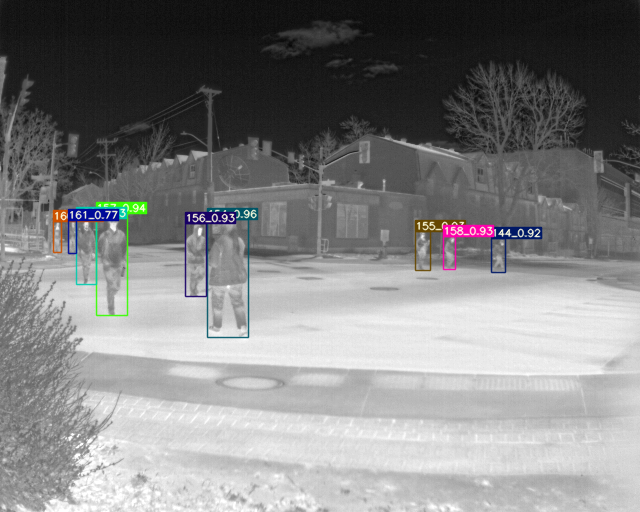} \\
				\multicolumn{6}{c}{(a) SORT}  \\
				\includegraphics[width=0.22\linewidth]{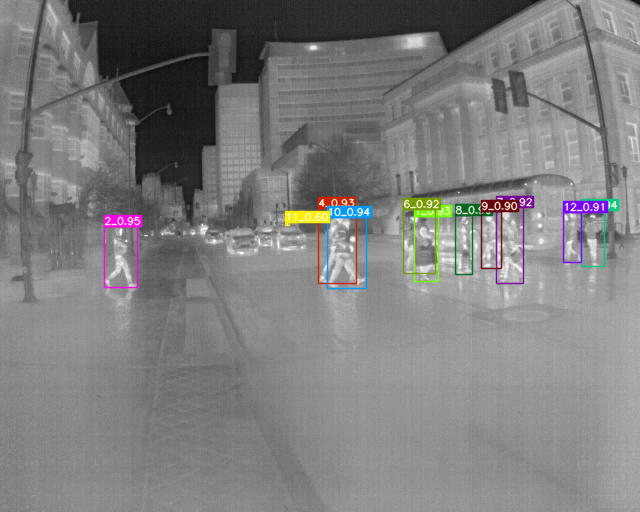} &
				\includegraphics[width=0.22\linewidth]{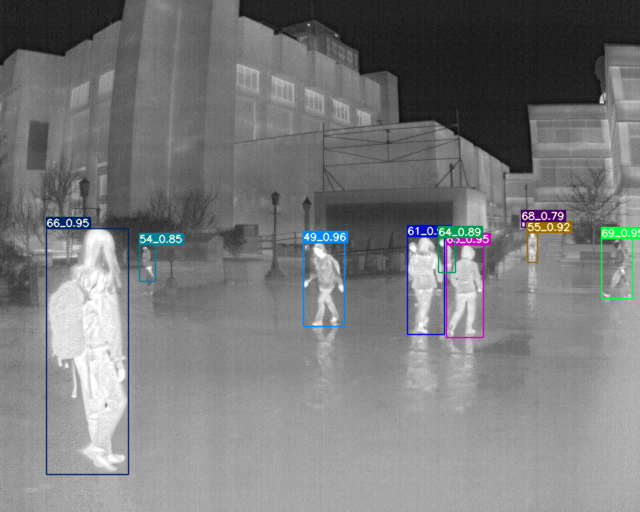} &
				\includegraphics[width=0.22\linewidth]{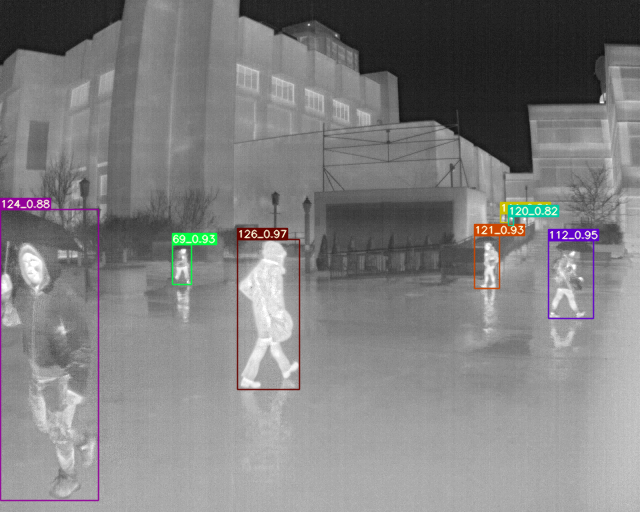} &
				\includegraphics[width=0.22\linewidth]{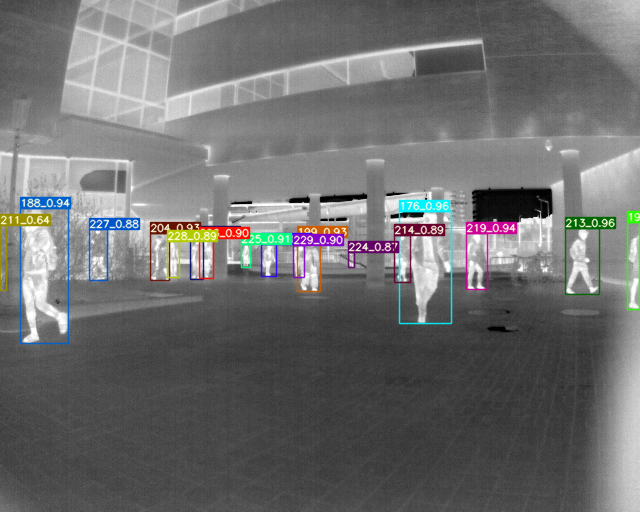} &
				\includegraphics[width=0.22\linewidth]{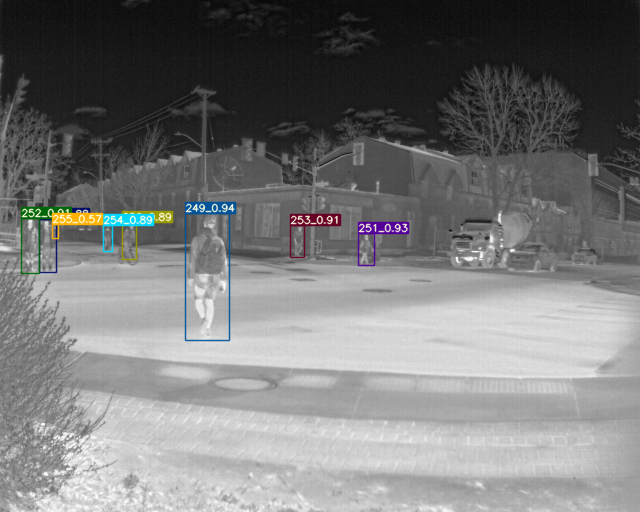} &
				\includegraphics[width=0.22\linewidth]{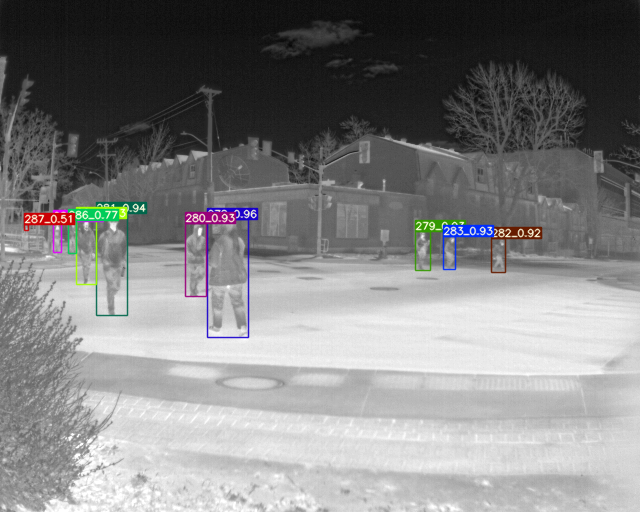} \\
				\multicolumn{6}{c}{(b) ByteTrack}  \\
				\includegraphics[width=0.22\linewidth]{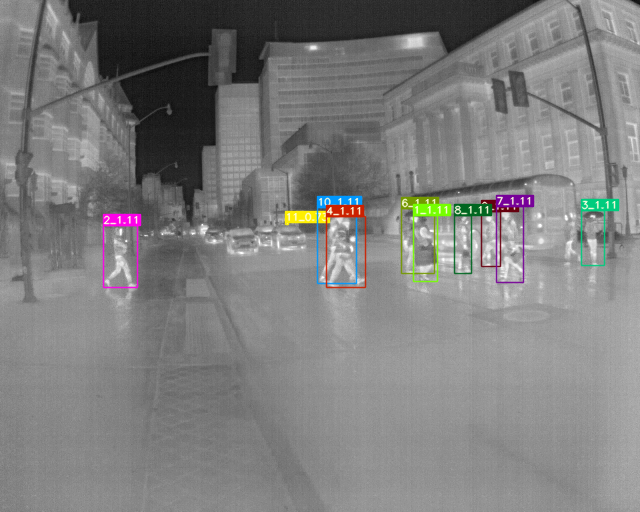} &
				\includegraphics[width=0.22\linewidth]{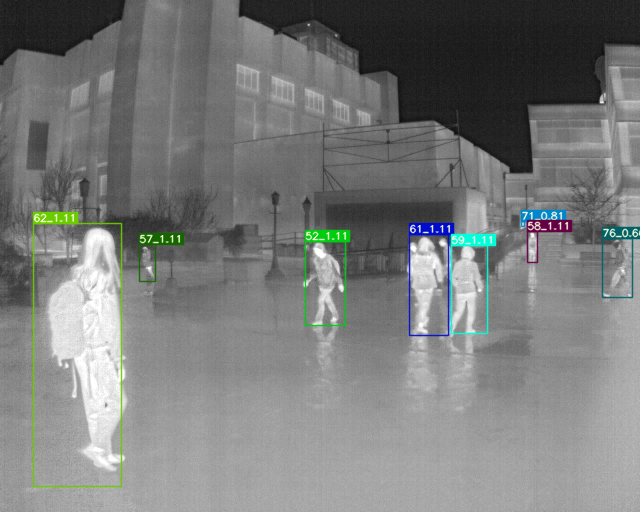} &
				\includegraphics[width=0.22\linewidth]{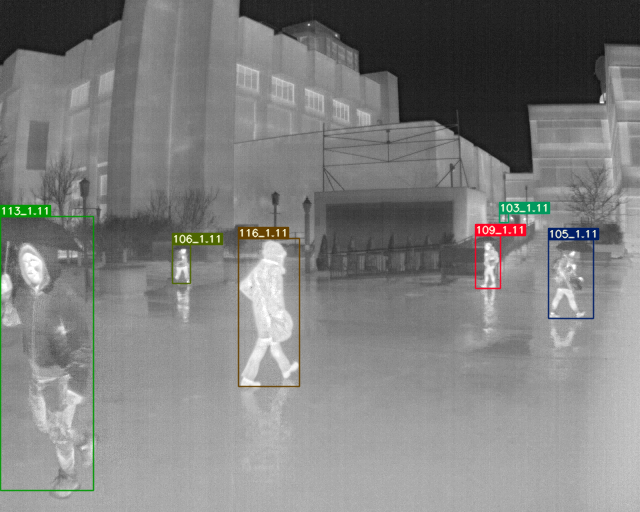} &
				\includegraphics[width=0.22\linewidth]{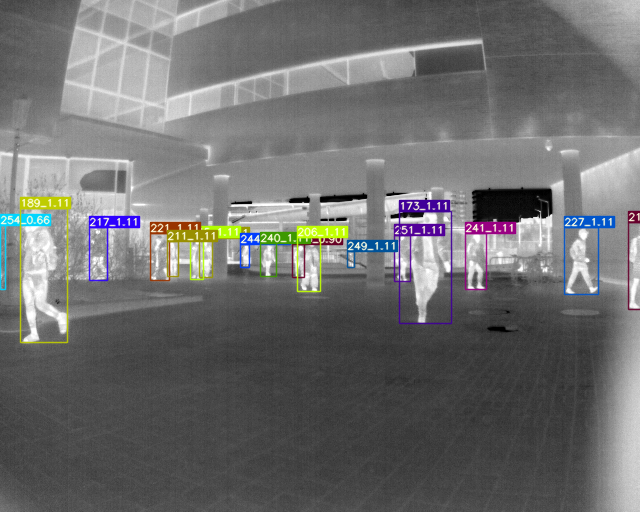} &
				\includegraphics[width=0.22\linewidth]{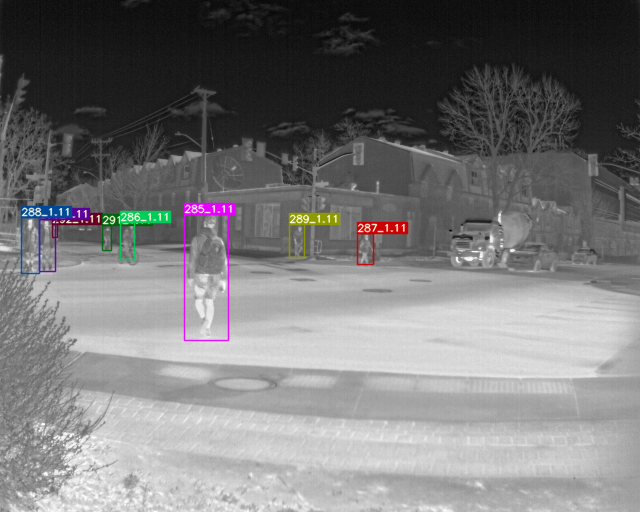} &
				\includegraphics[width=0.22\linewidth]{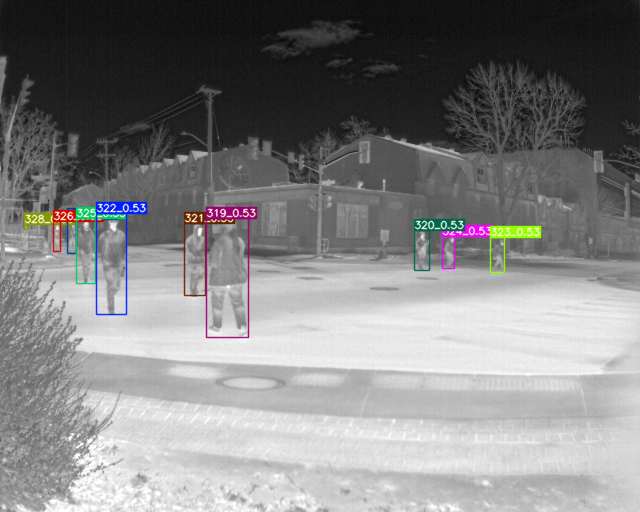} \\
				\multicolumn{6}{c}{(c) BoostTrack}  \\
				\includegraphics[width=0.22\linewidth]{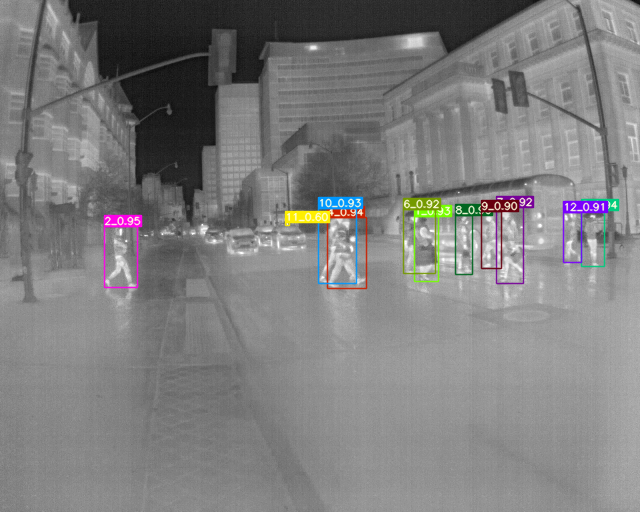} &
				\includegraphics[width=0.22\linewidth]{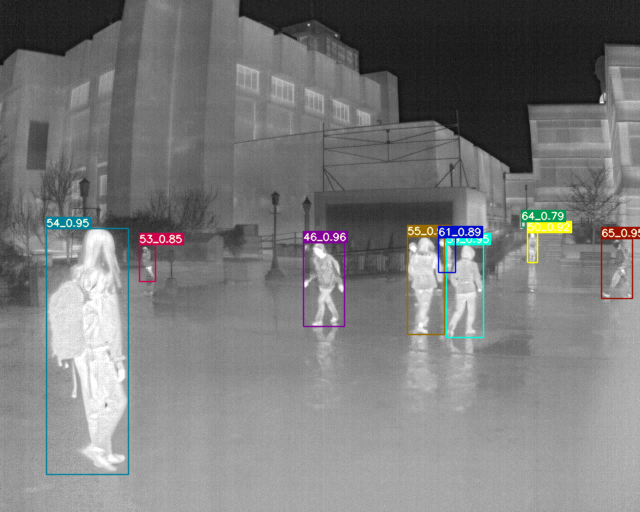} &
				\includegraphics[width=0.22\linewidth]{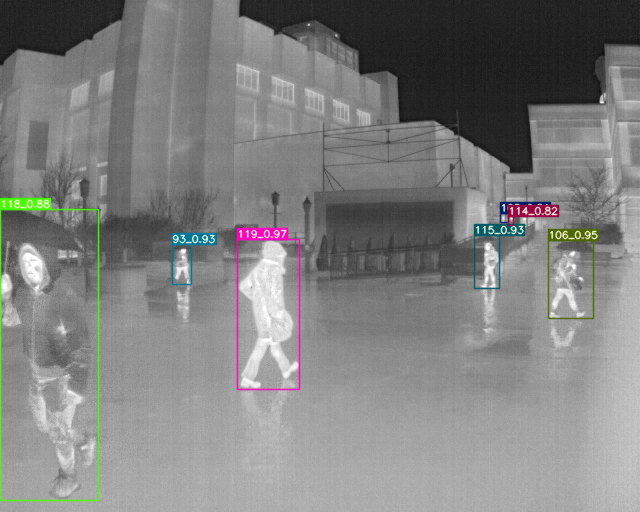} &
				\includegraphics[width=0.22\linewidth]{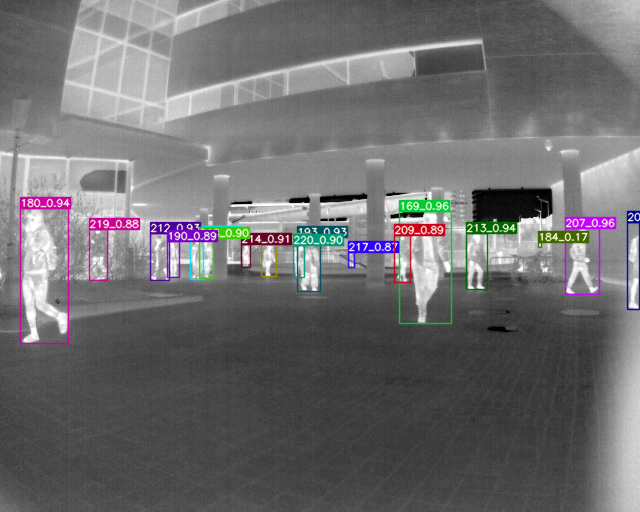} &
				\includegraphics[width=0.22\linewidth]{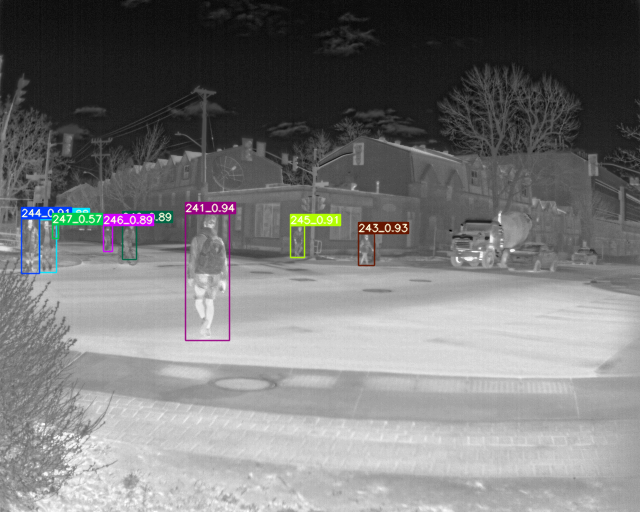} &
				\includegraphics[width=0.22\linewidth]{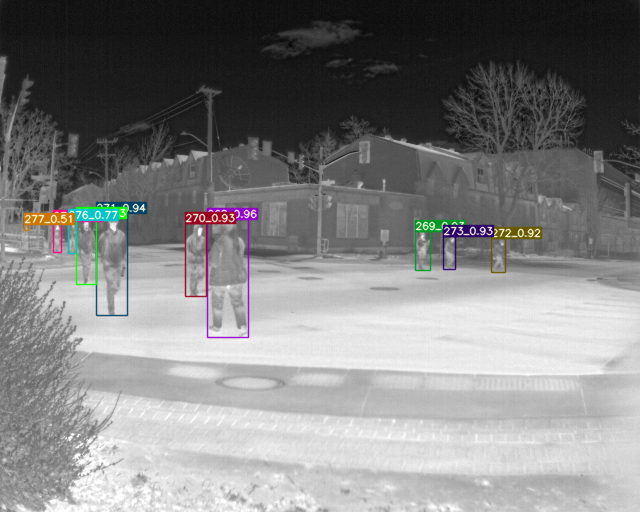} \\
				\multicolumn{6}{c}{(d) BoTSORT}  \\
				\includegraphics[width=0.22\linewidth]{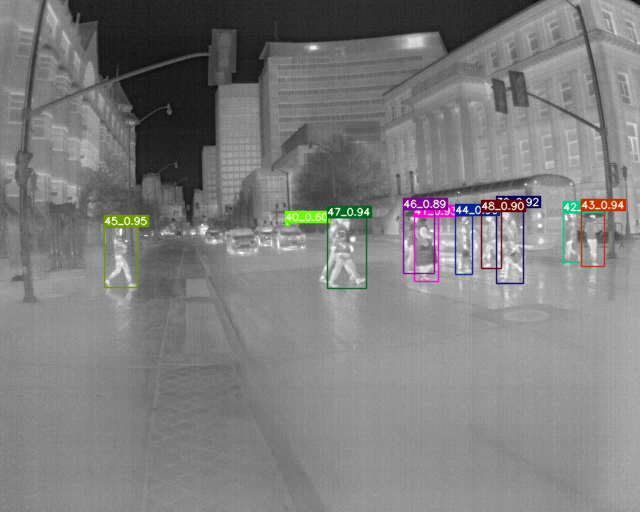} &
				\includegraphics[width=0.22\linewidth]{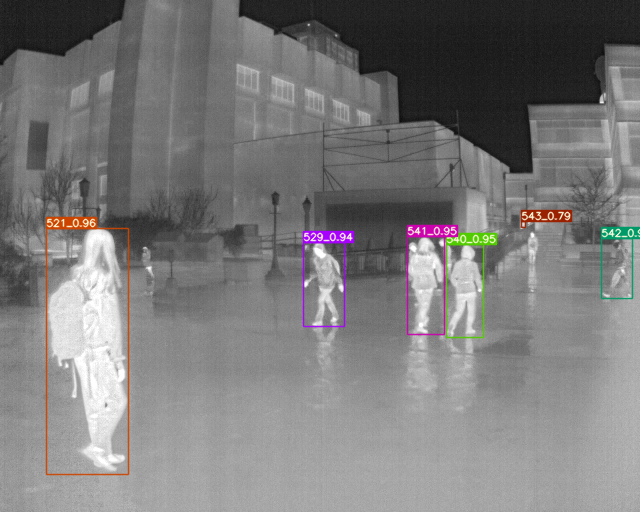} &
				\includegraphics[width=0.22\linewidth]{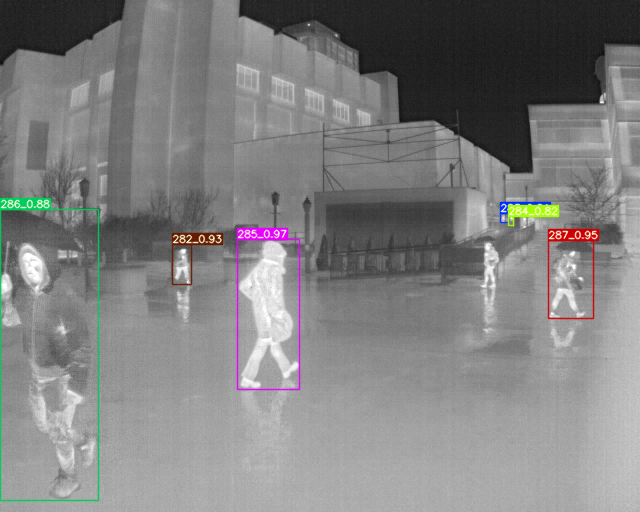} &
				\includegraphics[width=0.22\linewidth]{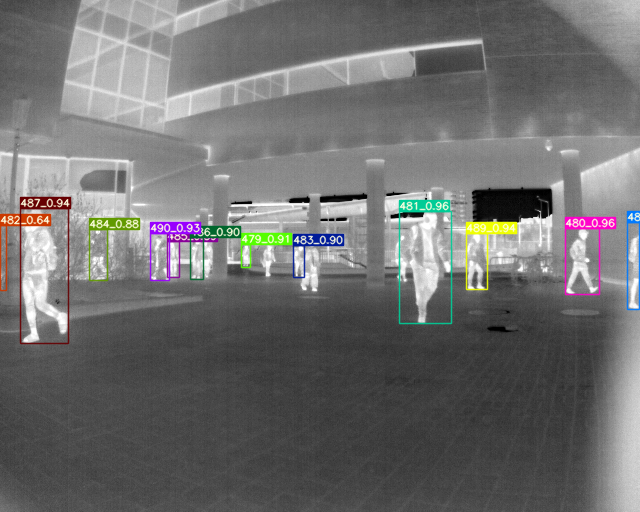} &
				\includegraphics[width=0.22\linewidth]{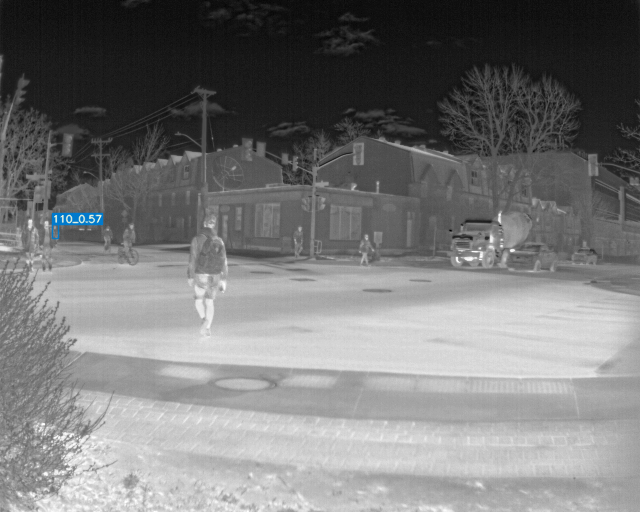} &
				\includegraphics[width=0.22\linewidth]{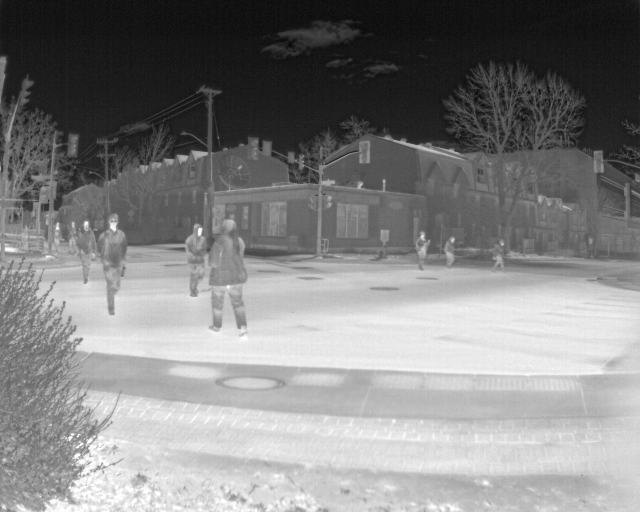} \\
				\multicolumn{6}{c}{(e) DiffMOT}  \\
			\end{tabular}
		}
		\hspace*{-0.89cm}
		\caption{The visualization of result of SORT, ByteTrack, BoostTrack, BoTSORT, and DiffMOT.}
		\label{fig:visualization_mot_result}
	\end{center}
\end{figure*}

\subsection{Evaluation Result}
\label{subsection:evaluation_result} 

As can be seen in Table \ref{tab:evaluation_result_server}, the evaluation table compares multi-object tracking (MOT) methods based on MOTA (tracking accuracy), MOTP (precision), and IDF1 (identity preservation). SORT achieves the highest MOTA (93.77) and IDF1 (80.78), making it the most accurate and consistent tracker. ByteTrack and BoTTrack perform similarly but with slightly lower scores, while BoostTrack and DiffMOT show weaker tracking accuracy, with MOTA around 86.3. MOTP scores indicate that SORT has the best localization precision (0.1366), while DiffMOT has the highest error (0.1611), suggesting lower bounding box accuracy. Overall, SORT is the most effective tracker, while ByteTrack and BoTTrack remain competitive, and BoostTrack and DiffMOT struggle with precision and tracking consistency

In addition, we implemented our methodology in the 1st Thermal Pedestrian Multiple Object Tracking Challenge (TP-MOT) \cite{El_Ahmar_2025_CVPR} at the Perception Beyond the Visible Spectrum workshop (PBVS) and achieved the highest ranking (as shown in Table \ref{tab:tmot_challenge}). 

\begin{table}
	\begin{center}	
		\resizebox{.3\textwidth}{!} {
		\begin{tabular}{@{}c|l|c@{}}
			\toprule
			Rank & Team        & Weighted Result \\
			\midrule
			\rowcolor[HTML]{D9D9D9}
			1    & AutoSKKU    & 0.71            \\
			2    & Fh-IOSB     & 0.55            \\
			3    & HNU-VPAI    & 0.42            \\
			4    & GyeongTiger & 0.29            \\
			5    & FFI BASED   & 0.25            \\
			\bottomrule
		\end{tabular}
	}
	\end{center}
	\vspace*{-0.3cm}
	\caption{Performance Metrics in TP-MOT.}
	\label{tab:tmot_challenge}
\end{table}

\section{Conclusion}
\label{section:conclusion}
The proposed hyperparameter tuning framework effectively enhances multi-object tracking (MOT) in thermal imagery by optimizing detection and object association in two key stages. By eliminating the need for complex re-identification models and ensuring real-time processing, the method improves tracking accuracy, robustness, and efficiency. This approach provides a lightweight and scalable solution for surveillance, security, and autonomous navigation applications in challenging thermal imaging conditions.

{
    \small
    \bibliographystyle{ieeenat_fullname}
    \bibliography{main}
}


\end{document}